\documentclass[authoryear,preprint,12pt]{elsarticle}

\usepackage{amssymb}
\usepackage{amsthm}
\usepackage{amsmath}
\usepackage{url}
\usepackage{threeparttable}
\usepackage[nolists,tablesfirst]{endfloat}
\DeclareMathOperator*{\argmax}{arg\,max}
\DeclareMathOperator*{\Var}{Var}

\journal{Expert Systems with Applications}

\begin{document}

\begin{frontmatter}

\title{Rasch-based high-dimensionality data reduction and class prediction with applications to microarray gene expression data}
\author{Andrej Kastrin\corref{cor1}}
\ead{andrej.kastrin@guest.arnes.si}
\author{Borut Peterlin\corref{cor2}}
\ead{borut.peterlin@guest.arnes.si}
\cortext[cor1]{Corresponding author. Tel.: +386 1 522 60 57; fax: +386 1 540 11 37}
\cortext[cor2]{Principal corresponding author. Tel.: +386 1 540 11 37; fax: +386 1 540 11 37}
\address{Institute of Medical Genetics, University Medical Centre Ljubljana, \v{S}lajmerjeva 3, SI-1000 Ljubljana, Slovenia}

\begin{abstract}
Class prediction is an important application of microarray gene expression data analysis. The high-dimensionality of microarray data, where number of genes (variables) is very large compared to the number of samples (observations), makes the application of many prediction techniques (e.g., logistic regression, discriminant analysis) difficult. An efficient way to solve this problem is by using dimension reduction statistical techniques. Increasingly used in psychology-related applications, Rasch model (RM) provides an appealing framework for handling high-dimensional microarray data. In this paper, we study the potential of RM-based modeling in dimensionality reduction with binarized microarray gene expression data and investigate its prediction accuracy in the context of class prediction using linear discriminant analysis. Two different publicly available microarray data sets are used to illustrate a general framework of the approach. Performance of the proposed method is assessed by re-randomization scheme using principal component analysis (PCA) as a benchmark method. Our results show that RM-based dimension reduction is as effective as PCA-based dimension reduction. The method is general and can be applied to the other high-dimensional data problems.
\end{abstract}


\end{frontmatter}


\section{Introduction}
\label{sec:introduction}
With decoding of the human genome and other eukaryotic organisms molecular biology has entered into a new era. High-throughput technologies, such as genomic microarrays can be used to measure the expression levels of essentially all the genes within an entire genome scale simultaneously in a single experiment and can provide information on gene functions and transcriptional networks \citep{Cordero2007}. The major challenge in microarray data analysis is due to their size, where the number of genes or variables ($p$) far exceeds the number of samples or observations ($n$), commonly known as the ``large $p$, small $n$'' problem. This takes it difficult or even impossible to apply class prediction methods (e.g., logistic regression, discriminant analysis) to the microarray data.

Class prediction is a crucial aspect of microarray studies and plays important role in the biological interpretation and clinical application of microarray data \citep{Chen2007, Larranaga2006}. For the last few years, microarray-based class prediction has been a major topic in applied statistics \citep{Slawski2008}. In a class prediction study, the task is to induce a class predictor (classifier) using available learning samples (i.e., gene expression profiles) from different diagnostic classes. Given the learning samples representing different classes, first the classifier is learned and then the classifier is used to predict the class membership (i.e., diagnostic class) of unseen samples \citep{Asyali2006}.

Generally, the performance of a classifier depends on three factors: the sample size, number of variables, and classifier complexity \citep{Jain2000, Raudys2006}. It was shown that for the fixed sample size, the prediction error of a designed classifier decreases and then increases as the number of variables grows. This paradox is referred to as the peaking phenomenon \citep{Hughes1968}. Moreover, some well known classifiers are even inapplicable in the setting of high-dimensional data. For example, the pooled within-class sample covariance in linear discriminant analysis (LDA) is singular if number of variables exceeds the number of samples. Similarly, in logistic regression the Hessian matrix will not have full rank and statistical packages will fail to produce reliable regression estimates \citep{Zhang2007}. Therefore, the number of samples must be larger than the number of variables for good prediction performance and appropriate use of classifiers. This naturally calls for the reduction of the ratio of sample size to dimensionality.

There are two major ways to handle high-dimensional microarray data in the class prediction framework. The first approach is to eliminate redundant or irrelevant genes (a.k.a. feature selection). The idea is to find genes with maximal discrimination performance and induce a classifier using those genes only \citep*{Asyali2006}. The most commonly used procedures of feature selection are based on simple statistical tests (e.g., fold change, $t$-test, ANOVA, etc.), which are calculated for all genes individually, and genes with the best scores are selected for classifier construction \citep{Dupuy2007, Jeffery2006}. The advantages of this approach are its simplicity, low computational cost, and interpretability. An alternative approach to overcome the problem of high-dimensionality is application of dimension reduction techniques (a.k.a. feature extraction). Generally, the aim of dimension reduction procedures is to summarize the original $p$-dimensional gene space in a form of a lower $K$-dimensional gene components space ($K<n$) that account for most of the variation in the original data \citep*{Jain2000}. Most commonly used methods for feature extraction with microarray gene expression data are principal component analysis (PCA) \citep{Alter2000, Chiaromonte2002, Holter2000}, partial least squares (PLS) \citep{Boulesteix2004, Boulesteix2007, Nguyen2002, Nguyen2004}, and sliced inverse regression (SIR) \citep{Antoniadis2003, Bura2003, Chiaromonte2002}. Although statistical analysis dealing with microarray data has been one of the most investigated areas in the last decade, there are only a few papers addressing the development and experimental validation of new methods and techniques for microarray dimension reduction. As \citet*{Fan2006} claimed, the high-dimensional data analysis will be one of the most important research topics in statistics in the nearest future. Here, we fill this gap by proposing a latent variable approach for handling high-dimensional microarray data and show its promising potential for class prediction.

\section{Background}
\label{sec:background}
The conceptual framework on latent variable modeling originates from psychometrics, starting at the beginning of the 20th century \citep{Fischer1995}. Utility of these models in biomedical research has only quite recently been recognized \citep{Li2001, Rabe-Hesketh2008}. By latent variable model we mean any statistical model that relates a set of observed variables to set of latent variables \citep{DeBoeck2004}. A latent variable is a variable that is not directly observable but does have a measurable impact on observable variables. Latent variables describe features that underlie the data. For example, a child's intelligence (i.e., latent variable) is typically assessed by measuring their answers to solving problems or items (i.e., observed variables) on intelligence test. The more items we ask of the child and the wider the breadth of items is included in the assessment, the more our understanding of that child's intellectual ability will be accurate.

The Rasch model (RM), originally proposed by \citet{Rasch1966}, is the simplest latent variable model. The idea behind the RM is that the probability of getting an item correct is a function of a latent trait or ability. For example, a child with higher intellectual ability would be more likely to correctly respond to a given item on an intelligence test. In psychological applications, data are usually given in a matrix, with rows being participants and columns being responses to a set of items. Microarray gene expression data can be represented in a similar way: columns are used to represent genes and rows are used to represent expression levels in biological samples. The RM can therefore be used to explain the observed gene expression patterns over different samples.

We assume that gene expression levels vary under the influence of $K$ latent gene factors. The idea behind our approach is to partition $p$ genes into $K$ functional subgroups and that covariations between genes with similar expression (i.e., genes in the same partition) could be described with a single gene factor. The factors in the model are latent, unobserved variables that account for the covariation among genes. It is assumed that genes with similar expression patterns might share biological function or might be under common regulatory control \citep{Do2008}. Moreover, it is particularly interesting to model a large set of genes as functions of fewer gene factors, because biologist believe the changes of mRNA levels are due to some regulatory factors \citep{Orlando2008}. Specifically, regulatory factors are proteins that bind certain DNA elements to regulate gene transcription to mRNA. Therefore, the gene factors obtained from latent modeling of gene expression could be interpreted as latent measurements of common regulatory factors of related genes. The number of gene factors is considered to be a meta-parameter and must be estimated or directly supplied based on researcher's prior knowledge. RM can then be used to estimate the magnitude of gene factors. Class prediction using standard prediction methods can then be carried out in the reduced space by using constructed gene factors as predictor variables.

The main objective of this paper is to evaluate the potential of RM-based dimensionality reduction with microarray gene expression data and investigate its prediction accuracy in the context of class prediction using LDA. To validate the proposed approach we apply a parallel PCA-based dimension reduction.

\section{Methods}
\label{sec:methods}
\begin{figure}[!h]
\includegraphics[width=\columnwidth]{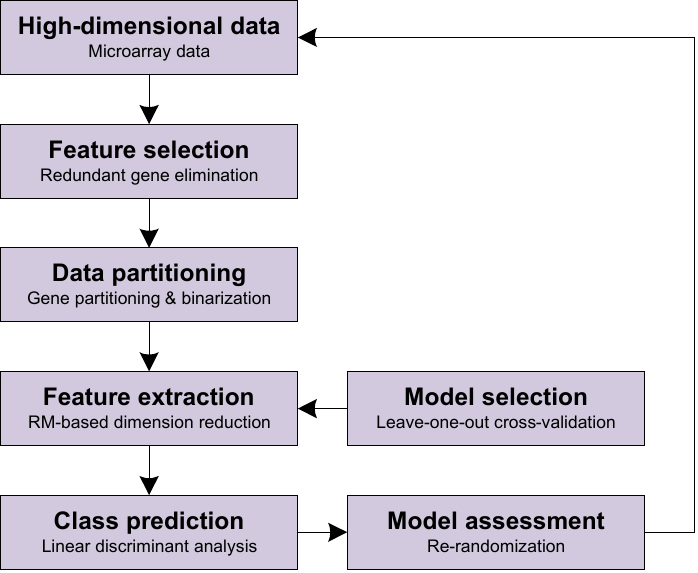}
\caption{The framework of dimension reduction.}
\label{fig:figure1}
\end{figure}

We propose a framework for dimension reduction and class prediction with application to gene expression data as illustrated in Figure~\ref{fig:figure1}. Our procedure consists of two basic steps: the first step is dimension reduction, in which data are reduced from high $p$-dimensional gene space to a lower $K$-dimensional gene factor space; the second step is class prediction, in which response classes are predicted using a class prediction method on the extracted gene factors.

\subsection{Data sets and preprocessing}
\label{subsec:data_sets}
We apply our algorithms to two publicly available microarray data sets which have been considered before by several authors. The Leukemia data set \citep{Golub1999} contains $n=72$ tissue samples with $p=7129$ genes: $47$ samples of acute lymphoblastic leukemia (ALL) and $25$ samples of acute myeloid leukemia (AML). The Prostate data set \citep{Singh2002} contains $n=102$ tissue samples with $p=12~600$ genes: $52$ prostate tumor samples and $50$ non-tumor prostate samples. Both data sets are from Affymetrix high-density oligonucleotide microarrays and are publicly available \citep{Dettling2004}.

For both data sets, the pre-processing steps are applied as follows \citep{Dudoit2002}: (a) thresholding, floor of $100$ and ceiling of $16~000$; (b) filtering, exclusion of genes with $\max/\min \leq 5$ and $(\max-\min) \leq 500$, where $\max$ and $\min$ refer to the maximum and minimum intensities for a particular gene across all samples; and (c) $\log_{10}$ transformation and standardization to zero mean and unit variance. The data were then summarized by a matrix $\mathbf{X} = (x_{ij})$, where $x_{ij}$ denotes the expression level for gene $j$ in sample $i$. The data for each sample consist of a gene expression profile $\mathbf{x}_{i}=\left(x_{1i},\ldots,x_{pi}\right)^{T}$ and a class label $y_{i}$. After data preprocessing the dimension of the matrix $\mathbf{X}$ was $72 \times 3571$ and $102 \times 6033$ for Leukemia and Prostate data set, respectively.

\subsection{Feature selection}
\label{subsec:feature_selection}
Although the procedure described here can handle a large number (thousands) of genes, the number of genes may still be too large for practical use. The model assessment procedure is very CPU-expensive and therefore time-consuming process, because it requires fitting the data many times due to cross-validation and re-randomization. Furthermore, a considerable percentage of the genes do not show differential expression across groups and only a subset of genes is of interest.

We used two different methods for feature selection in this study. First we performed an unsupervised random subset selection, consisting of $p^{\ast}$ ($p^{\ast}<p$) genes from the set of all genes, as described by \citet{Dai2006}. We selected $p^{\ast}$ genes with $p^{\ast} = \lbrace 50,100,200 \rbrace$ from both experimental data sets.

As the supervised alternative, we applied Welch's $t$-test \citep*{Jeffery2006} to embed the class information in feature selection process and thus improved prediction accuracy. Welch's $t$-test provides the measure of the statistical significance of changes in gene expression between classes. Welch's $t$-test defines the statistic $t$ by
\begin{equation*}
t = \frac{\overline{x}_{1}-\overline{x}_{2}}{\sqrt{s^{2}_{1} / n_{1} + s^{2}_{2} / n_{2}}},
\end{equation*}
where $\overline{x}_{g}$, $s^{2}_{g}$, and $n_{g}$ are the mean, sample variance, and sample size of the class $g$ ($g=1,2$) for each gene, respectively. Feature selection was carried out based on absolute value of the $t$-statistic and the top $p^{\ast}$ genes with $p^{\ast} = \lbrace 50,100,200 \rbrace$ were used for further processing.

\subsection{Feature extraction}
\label{subsec:feature_extraction}
Here we describe the core of feature extraction based on the RMs and then briefly outline the PCA method, which we used as a benchmark.
\subsubsection{Rasch model}
\label{subsubsec:rasch}
In this subsection we first give a short overview of the RM theory in its original form, and then present its application to gene expression data.
 
\begin{figure}[!h]
\includegraphics[width=\columnwidth]{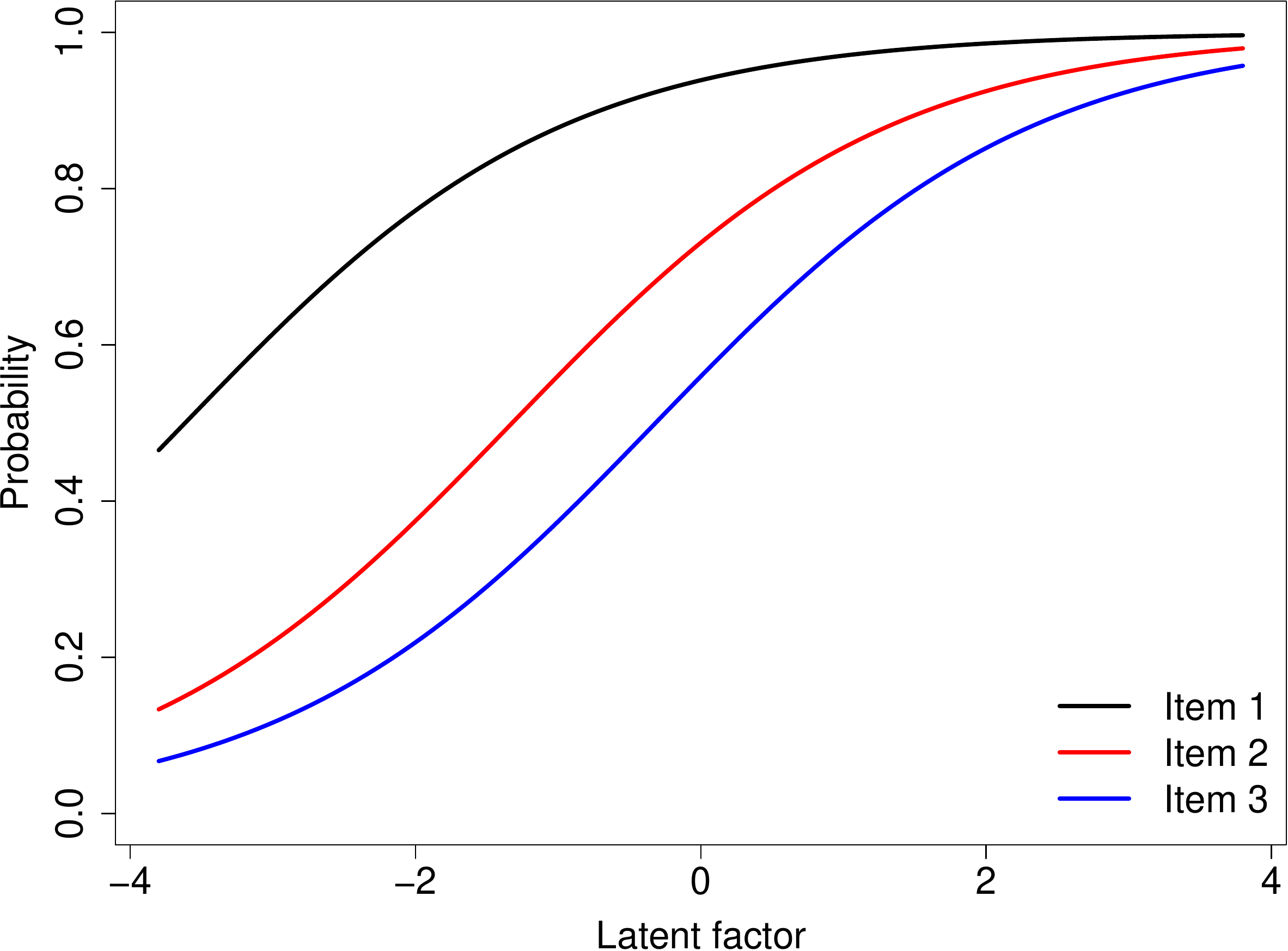}
\caption{Rasch probabilities as a function of the value of the latent factor for three different items. The person is likely to respond correctly to the Item 1 and unlikely to respond correctly to Item 3. Item parameters are $\beta_{1}=-3.62$, $\beta_{2}=-1.32$, and $\beta_{3}=-0.32$, respectively.}
\label{fig:figure2}
\end{figure}

The RM is a simple latent factor model, primarily used for analyzing data from assessments to measure psychological constructs such as personality traits, abilities, and attitudes \citep{Fischer1995}. Assume that we have $I$ persons and $J$ items. Let $y_{ij}$ be the response of person $i$ to the item $j$, where the $y_{ij}$ is `1' if person $i$ answered item $j$ correctly and `0' otherwise. In the RM, the probability of the outcome $y_{ij}=1$ is given by
\begin{equation*}
P\left(y_{ij}=1 \vert \eta_i\right) = \frac{\exp\left(\beta_{j} + \eta_{i}\right)}{1+\exp\left(\beta_{j}+\eta_{i}\right)},
\end{equation*}
for $i=1, \ldots, I$ and $j=1, \ldots, J$. $\eta_{i}$ is the person parameter that denotes the latent factor of the $i$th person that is measured by the item $j$ and $\beta_{j}$ is the item parameter, which denotes the difficulty of the item $j$. Difficulty of the item describes the region of the latent trait distribution where the probability of producing a specific response changes from low to high. Probability of the response is monotonous in both person and item parameters. Figure~\ref{fig:figure2} plots the Rasch probabilities as a function of the value of the latent factor ($\eta$) for three different items. It can be seen that for a given item, persons with larger $\eta$ value tend to have greater probability of expressing high scores on the latent factor, and for a given person, the response probabilities are different for items with different $\beta$ values.

The final step of the RM-based modeling is to derive latent scores from the item responses. Latent score is the total score of person $i$ over $J$ items. The most common approach to calculate latent scores is to use the expectation of the posterior distribution of $\eta_{i}$ given $\mathbf{y}_{i}$ with parameter estimates plugged in \citep{Rabe-Hesketh2008}. Details on the calculation can be found in \citet*{Fischer1995}.

\paragraph{Application of Rasch model to gene expression data}
In terminology of RM, we denote each gene as an ``item'' and each sample as a ``person''. The expression level $x_{ij}$ of gene $j$ in sample $i$ is the response of a given sample to a given gene. Our main assumption is, that one-dimensional latent model may not hold for the complete set of $p^{\ast}$ genes selected in the gene filtering step (see Subsection~\ref{subsec:feature_selection}). Based on the assumption that expressional similarity implies functional similarity of the genes (and vice versa), we assume that genes with similar expression patterns determine one latent factor \citep{Do2008}.

To identify coexpressed genes we used $k$-means clustering \citep{Gan2007} to partition $p^{\ast}$ genes into $K$ partitions ($k=1, \ldots, K$) based on their gene expression profiles over $n$ samples. $K$-means clustering is a simple and widely used partitioning algorithm. Its helpfulness in discovering groups of coexpressed genes has been demonstrated \citep{Richards2008}. The optimal number of $K$ is estimated following the procedure described in Subsection~\ref{subsec:model_selection}.

To apply the RM to the gene expression data, we need to discretize the gene expression data matrix $\mathbf{X}$ into binary form. We use the median as a cut-off point for discretization. The intensity of every gene expression value is compared with the median gene expression data of the $\mathbf{X}$ and assigned a `$1$' if it is above and `$0$' otherwise. Note that gene partitioning is done before discretization step.

We fit a RM for genes in each of the $K$ partitions respectively and calculate gene factor scores. Specifically, we construct $K$ latent gene factors on which each gene in the $k$ partition is located. The measure for each sample for each gene factor is then estimated. To fit the RM for genes in the $k$th partition, let $i$ be the sample index, and $j$ be the gene index, for $i=1, \ldots,n$, and $j=1,\ldots,p_k$, and let $\eta_i=\eta_{ik}$ be the latent gene factor for the $i$th sample which is determined by the genes in the $k$th partition, and $\beta_{j}$ be the gene specific parameter for the $j$th gene. Class prediction is then carried out in the reduced space by using the gene factors.

\subsubsection{Principal component analysis}
\label{subsubsec:pca}
PCA is the most commonly used technique for dimension reduction in microarray data analysis \citep*{Alter2000}. The main idea behind the PCA is to reduce the dimensionality of a data set, while retaining as much as possible the variation in the original variables \citep{Hastie2001}. This is achieved by transforming the $p^{\ast}$ original variables $\mathbf{X} = [\mathbf{x}_{1}, \ldots, \mathbf{x}_{p}]$ to a new set of $K$ predictor variables $\mathbf{T} = [\mathbf{t}_{1}, \ldots, \mathbf{t}_{K}]$, which are linear combinations of the original variables. More formally, PCA sequentially maximizes the variance of a linear combination of the original variables,
\begin{equation*}
\mathbf{a}_K = \argmax_{\mathbf{a}^{T}\mathbf{a}=1} \Var \left(\mathbf{Xa}\right)
\end{equation*}
subject to the constraint $\mathbf{a}_{i}^{T} \mathbf{S_{X}} \mathbf{a}_{j}$, for all $1 \leq i < j$, where $\mathbf{S_{X}}$ is covariance matrix of the original data. The orthogonal constraint ensures that the linear combinations are uncorrelated. Linear combinations $\mathbf{t}_{i} = \mathbf{Xa}_{i}$ are known as the principal components. These linear combinations represent the selection of a new coordinate system obtained by rotating the original system. The new axes represent the directions with maximum variability and are ordered in terms of the amount of variation of the original data they account for. The first principal component accounts for as much of the variability in the original data as possible, and each succeeding component accounts for as much of the remaining variability as possible. Computations of the weighting vectors $\mathbf{a}$ involves the calculation of the eigenvalue decomposition of a data covariance matrix $\mathbf{S_X}$,
\begin{equation*}
\mathbf{S_{X}}\mathbf{a}_{i} = \lambda_{i}\mathbf{a}_{i}
\end{equation*}
where $\lambda_{i}$ is the $i$th eigenvalue in the descending order for $i = 1, \ldots, K$ and $\mathbf{a}_{i}$ is the corresponding eigenvector. The eigenvalue $\lambda_{i}$ measures the variance of the $i$th principal component and the eigenvector $\mathbf{a}_{i}$ provides the loadings for the linear transformation. The number of components $K$ is specified on the basis of researcher's prior knowledge or determined using dedicated procedures (e.g., Kaiser-Guttman rule, Cattell's scree test, etc.). Class prediction using standard methods can then be carried out in the reduced space by using the constructed principal components.

\subsection{Class prediction}
\label{subsec:class_prediction}
After dimension reduction, the high dimension of $p^{\ast}$ is now reduced to a lower dimension. The original data matrix is approximated by matrix of gene factors ($n \times K$, where $K < n$), constructed by RMs or PCA, as described in the previous section. To avoid confusion, we use the term ``factor'' to refer to both latent factor obtained from RM analysis and principal component derived from PCA. Once the $K$ gene factors are constructed we consider prediction of the response classes.

To describe the class prediction problem formally, let we have a learning set $\mathcal{L}$ consisting of samples whose class is known and a test set $\mathcal{T}$ consisting of samples whose class has to be predicted. Denote the data matrix corresponding to $\mathcal{L}$ as the learning data matrix $\mathbf{X}_{L}$, and the data matrix corresponding to $\mathcal{T}$ as the test data matrix $\mathbf{X}_{T}$. The vector containing the classes of the samples from $\mathcal{L}$ is denoted as $\mathbf{y}_{L}$. The goal is to build a rule implementing the information from $\mathbf{X}_{L}$ and $\mathbf{y}_{L}$ in order to predict the class $g$ of the $i$th sample from the test set given the gene expression profile $\mathbf{x}_{new,i}$:
\begin{alignat*}{2}
\delta \left(.,\mathbf{X}_{L},\mathbf{y}_{L} \right): & \;\; \mathbb{R}^{K} \;\; & \rightarrow & \;\; \lbrace 1,\ldots, G \rbrace\\
& \;\; \mathbf{x}_{new,i} \;\; & \mapsto & \;\; \delta \left(\mathbf{x}_{new,i},\mathbf{X}_{L},\mathbf{y}_{L} \right).
\end{alignat*}

Because our focus here is on dimension reduction, we fixed the class prediction step with LDA, although other methodologies can be used \citep*{Bellazzi2008, Dudoit2002}. A short description of the LDA method is given in the following \citep{Boulesteix2004}. Suppose we have $K$ predictor variables. The random vector $\mathbf{x}=(X_{1},\ldots,X_{K})^{T}$ is assumed to a multivariate normal distribution within class $g=1, \ldots, G$ (in our procedure $G=2$) with mean $\boldsymbol{\mu}_{g}$ and covariance matrix $\boldsymbol{\Sigma}_{g}$. In LDA, $\boldsymbol{\Sigma}_{g}$ is assumed to be the same for all classes: for all $g$, $\boldsymbol{\Sigma}_{g} = \boldsymbol{\Sigma}$. Using estimates $\hat{\boldsymbol{\mu}}_{g}$ and $\hat{\boldsymbol{\Sigma}}$ in place of $\boldsymbol{\mu}_{g}$ and $\boldsymbol{\Sigma}$, the discriminant rule assign the $i$th new observation $\mathbf{x}_{new,i}$ to the class
\begin{equation*}
\delta\left(\mathbf{x}_{new,i}\right) = \argmax_{g}\left(\mathbf{x}_{new,i} - \hat{\boldsymbol{\mu}}_{g}\right) \, \hat{\boldsymbol{\Sigma}}^{-1} \, (\mathbf{x}_{new,i} - \hat{\boldsymbol{\mu}}_{g})^{T}.
\end{equation*}

LDA has been well studied and widely used for class prediction problems. LDA relies on a hypothesis of multinormality, and assumes that the classes have the same covariance matrix. Although these hypotheses are rarely satisfied with real data sets, LDA generally gives good results. Studies have demonstrated favorable prediction performances of LDA models when compared with more complicated and computationally intensive algorithms such as neural networks and tree method \citep*{Dudoit2002, Tibshirani2002}. For the details of the calculation we refer the reader to \citet{Ripley1996}.

\subsection{Model selection}
\label{subsec:model_selection}
The number of gene factors $K$ is a meta-parameter in the procedure. We estimate $K$ on the learning set $\mathcal{L}$ using leave-one-out cross-validation (LOOCV). LOOCV has been shown to give an almost unbiased estimator of the prediction error \citep*{Hastie2001}, and therefore provide a sensible criterion for our purposes. In a nutshell, a subset of $p^{\ast}$ genes is selected ($p^{\ast}<p$) from $\mathcal{L}$, and one of the samples is left-out. The feature extraction models are fitted to all but the left-out sample (see Subsection~\ref{subsec:feature_extraction}). The fitted models are then used to predict the class of the left-out sample (see Subsection~\ref{subsec:class_prediction}). This is repeated for all samples in the learning data set $n_{L}$ with $K$ taking successively different values. The mean error rate (\textit{MER}) over the $n_{L}$ runs is computed for each value of $K$ by
\begin{equation*}
\mathrm{\textit{MER}} = \frac{1}{n_{L}}\sum\limits_{i=1}^{n_{L}}I(\hat{y}_{i} \neq y_{i}),
\end{equation*}
where $\hat{y}_{i}$ is the predicted response class and $y_{i}$ is the observed response class. $I$ is the indicator function ($I(A)=1$ if $A$ is true, $I(A)=0$ otherwise). The value of $K$ minimizing the \textit{MER} is selected and denoted as $K^{\ast}$. We select $K = \lbrace 1,2,3,4,5 \rbrace$ in our experiments.

\subsection{Performance evaluation}
\label{subsec:evaluation}
To assess the performance of the RM- and PCA-based dimension reduction methods in the framework of class prediction we perform a re-randomization study. This evaluation approach was first used by \citet*{Dudoit2002}. The reader may refer to the review of \citet{Boulesteix2008} on this subject. The procedure consists of the six steps as follows.
\begin{enumerate}[Step 1.]
\item For each data set, create $R=100$ random partitions into learning data set $\mathcal{L}$ with $n_{L}$ samples and a test set $\mathcal{T}$ with $n_{T}$ samples ($n_{L} + n_{T} = n$). Denote $\mathbf{X}_{L}$ as the learning data matrix of size $n_{L} \times p$, and $\mathbf{X}_{T}$ as the test data matrix of size $n_{T} \times p$.
\item Select a subset of $p^{\ast}$ genes from the set of all genes $p$ from matrix $\mathbf{X}_{L}$ using one of the gene selection methods, resulting in $\mathbf{X}^{\ast}_{L}$ matrix of size $n_{L} \times p^{\ast}$ and $\mathbf{X}^{\ast}_{T}$ matrix of size $n_{T} \times p^{\ast}$ (see Subsection~\ref{subsec:feature_selection}). 
\item Use the learning data matrix $\mathbf{X}^{\ast}_{L}$ to determine the number of latent factors $K^{\ast}$, by LOOCV (see Subsection~\ref{subsec:model_selection}).
\item Perform dimension reduction using RM- or PCA-based dimension reduction (see Subsection~\ref{subsec:feature_extraction}). Let $\mathbf{W}$ denote the matrix containing the factor loadings of size $p^{\ast} \times K^{\ast}$. Compute the matrix $\mathbf{Z}_{L}$ of gene factors for the learning data set ($\mathbf{Z}_{L}=\mathbf{X}^{\ast}_{L} \times \mathbf{W}$), and the matrix $\mathbf{Z}_{T}$ of gene factors for the test data set ($\mathbf{Z}_{T}=\mathbf{X}^{\ast}_{T} \times \mathbf{W}$).\footnote{Note that matrix $\mathbf{W}$ refers to component loadings, and matrix $\mathbf{Z}$ to gene components when PCA is performed.}
\item Fit the class prediction model to the learning gene factors $\mathbf{Z}_{L}$. Predict the classes of samples in the test set using the fitted classifier and the test gene factors, $\mathbf{Z}_{T}$ (see Subsection~\ref{subsec:class_prediction}).
\item Repeat all above steps $R$ times with re-randomization of the whole data set. The mean error rate ($\mathrm{\textit{MER}}$) for each method is given by
\begin{equation*}
\mathrm{\textit{MER}} = \frac{1}{R}\sum\limits_{r=1}^{R} \frac{1}{n_{T}}\sum\limits_{i=1}^{n_{T}}I(\hat{y}_{i} \neq y_{i}),
\end{equation*}
where $\hat{y}_{i}$ is the predicted response class and $y_{i}$ is the observed response class. $I$ is the indicator function ($I(A)=1$ if $A$ is true, $I(A)=0$ otherwise).
\end{enumerate}
Although the $\mathrm{\textit{MER}}$ is the most widely used metric for measuring the performance of the prediction systems, it considers all mispredictions as equally damaging \citep*{Boulesteix2008}. It has been demonstrated \citep{Huang2005} that, when the prior class probabilities are different, this measure is not appropriate because it does not consider misprediction cost, is strongly biased to favor the majority class, and is sensitive to class skewness. To overcome these problems the performance evaluation was also carried out via receiver operator characteristic (ROC) analysis. For comprehensive introduction to ROC analysis we refer the reader to \citet{Fawcett2006}. The ROC analysis was performed by plotting true positive rate (sensitivity) versus the false positive rate (1$-$specificity) at various threshold values, and the resulting curve was integrated to give an area under the curve (\textit{AUC}) value. \textit{AUC} is a measure of the discriminative power of the classes using the given features and classifier, and varies from $\mathrm{\textit{AUC}} = 0.5$ for non-distinguishable classes to $\mathrm{\textit{AUC}} = 1.0$ for perfectly distinguishable classes \citep{Huang2005}. The \textit{AUC} can be interpreted as the probability that two random samples from the two classes will be predicted correctly, and is invariant to changes in class proportions (unlike \textit{MER}). An $\mathrm{\textit{AUC}} \geq 0.7$ is generally considered acceptable, $\mathrm{\textit{AUC}} \geq 0.8$ as good, and $\mathrm{\textit{AUC}} \geq 0.9$ as excellent prediction performance.

\subsection{Software}
\label{subsec:software}
All computations were carried out in the $\mathsf{R}$ software environment for statistical computing and graphics \citep{R}. $K$-means clustering was performed using generic \verb+kmeans+ function. Binarization of continuous gene expression values was performed by \verb+binarize+ function in the \verb+minet+ package. The function \verb+generate.split+ of the \verb+WilcoxCV+ package was used to generate random splitting into learning and test data sets. RM analysis was performed using \verb+ltm+ package. PCA was conducted using generic \verb+prcomp+ function. LDA was carried out using \verb+lda+ function in the \verb+MASS+ package. ROC analysis was performed using the \verb+caret+ package. The procedures described here can be reproduced using the $\mathsf{R}$ scripts available from \url{http://www2.arnes.si/~akastr1/rasch/}

\section{Results}
\label{sec:results}
We illustrate the interest of RM-based dimension reduction by considering applications for the class prediction of microarray data. We compare the results from our procedure with the performance of the PCA-based approach. We will consider in turn the Leukemia and Prostate data sets, as described previously in Subsection~\ref{subsec:data_sets}

\subsection{Application to Leukemia data set}
\label{subsec:leukemia}
After data preprocessing, we applied the proposed performance evaluation procedure on the Leukemia data set. First, we consider $p^{\ast} = \lbrace 50,100,200 \rbrace$ randomly selected genes and used $R=100$ random subsets. We randomly split each subset of genes into two data sets: a learning set with $n_{L}=36$ samples and a test set with $n_{T}=36$ samples. We used LOOCV procedure on the learning set to determine the number of gene factors (components in the case of PCA), and the test set for evaluating prediction performances. In total, $3600$ class predictions were calculated using each of the dimension reduction method based on $100$ randomization trials.

\begin{figure}[!h]
\includegraphics[width=\textwidth]{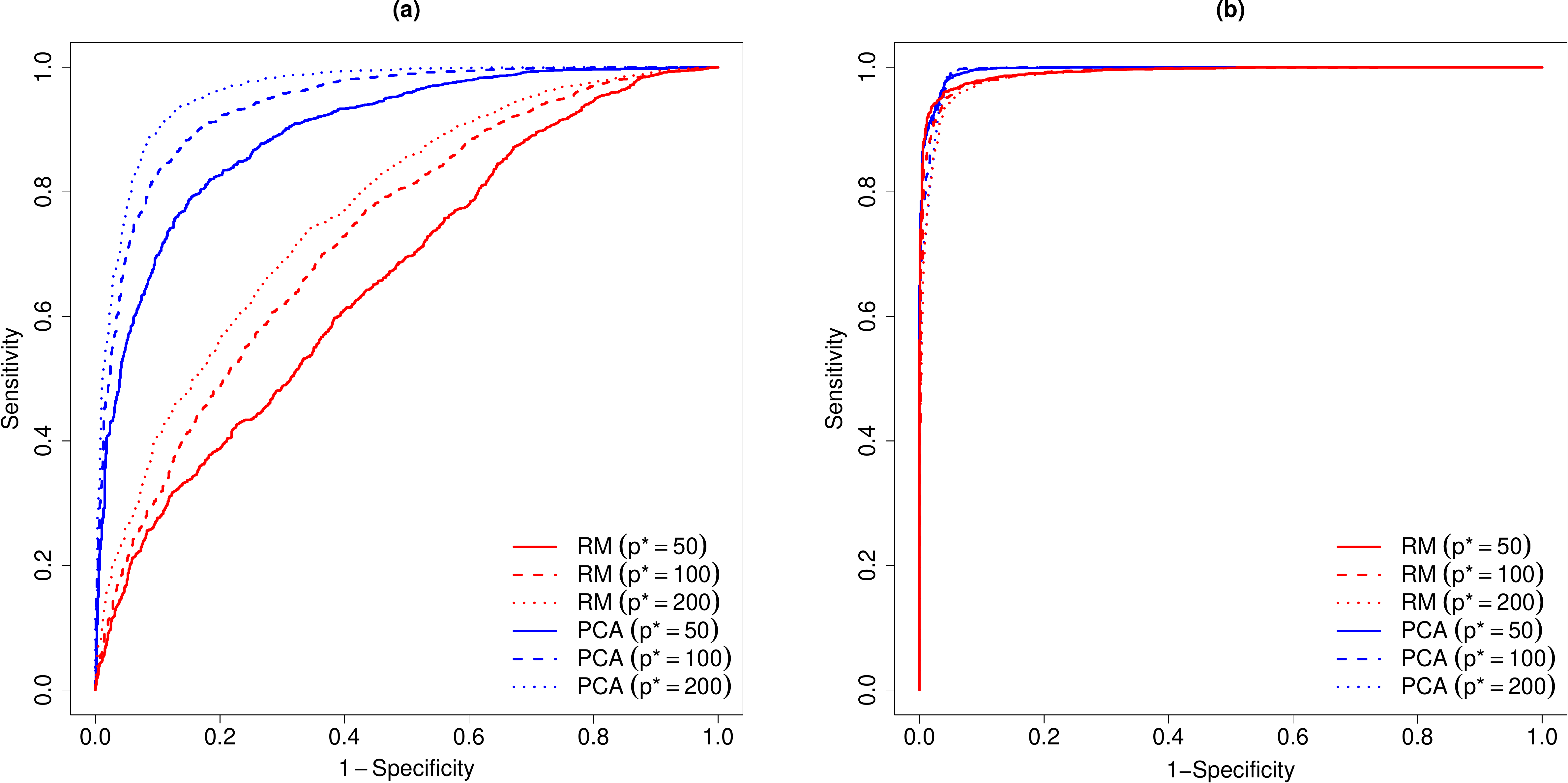}
\caption{Receiver operating characteristic curves (ROC) for RM- and PCA-based prediction models using random (a) and supervised (b) gene selection on the Leukemia data set with $36/36$ split of samples over $100$ randomization trials.}
\label{fig:figure3}
\end{figure}

Table~\ref{tab:table1} gives the estimated mean error rates (\textit{MER}), average values of the estimated meta-parameters ($K^{\ast}$), corresponding standard deviations, and areas under the ROC curves (\textit{AUC}) with the considering methods, for different number of variables. A ROC curve analysis is depicted in Figure~\ref{fig:figure3}(a). It can be seen from Table~\ref{tab:table1} that \textit{MER} decreases with the increase of the size of gene subsets $p^{\ast}$. Inversely, the \textit{AUC} increases when more predictor genes are included in model building. At any of the subsets size, the \textit{MER}s of PCA-based class prediction are lower that that of RM-based procedure. \textit{AUC} scores suggest acceptable prediction performance for RM-based dimension reduction model and excellent performance for the PCA-based model.

\begin{table}[!h]
\begin{threeparttable}
\caption{Prediction performances of the RM- and PCA-based prediction models using random gene selection on the Leukemia data set with $36/36$ split of samples over $100$ randomization trials.}
\label{tab:table1}
\begin{tabular}{rccccccc}
\hline
& \multicolumn{3}{c}{RM-LDA} & & \multicolumn{3}{c}{PCA-LDA} \\
\cline{2-4}
\cline{6-8}
\multicolumn{1}{c}{$p^{\ast}$} & \textit{MER} & $K^{\ast}$ & \textit{AUC} & & \textit{MER} & $K^{\ast}$ & \textit{AUC} \\
\hline
$50$ & $0.31\,(0.10)$ & $2.05\,(1.23)$ & $0.66$ & & $0.17\,(0.09)$ & $3.20\,(1.29)$ & $0.90$ \\
$100$ & $0.29\,(0.10)$ & $3.06\,(1.42)$ & $0.73$ & & $0.12\,(0.07)$ & $3.26\,(1.14)$ & $0.94$ \\
$200$ & $0.27\,(0.10)$ & $3.44\,(1.44)$ & $0.77$ & & $0.09\,(0.06)$ & $3.23\,(1.17)$ & $0.96$ \\
\hline
\end{tabular}
\begin{tablenotes}[flushleft]
\footnotesize
\setlength{\itemindent}{-3pt}
\item Legend: RM-LDA -- RM-based class prediction; PCA-LDA -- PCA-based class prediction; $p^{\ast}$ -- number of selected genes; \textit{MER} -- mean error rate; $K^{\ast}$ -- estimated number of gene factors (components); \textit{AUC} -- area under the ROC curve.
\end{tablenotes}
\end{threeparttable}
\end{table}

Next, we present the results of performance evaluations using subsets of genes selected based on Welch's $t$-test. As described in Subsection~\ref{subsec:feature_selection}, genes were ranked according to absolute value of the $t$-statistic and the top $p^{\ast}$ genes with $p^{\ast} = \lbrace 50,100,200 \rbrace$ were used for extraction of gene factors. The analysis design was the same as for the randomly selected subset of genes described previously. The performance results on both methods are given in Table~\ref{tab:table2}. Corresponding ROC curves are presented in Figure~\ref{fig:figure3}(b). Comparing the results in Table~\ref{tab:table2} with the results given in Table~\ref{tab:table1}, one can see that the accuracy of class prediction has been improved significantly. \textit{MER}s of all two methods are reduced. Regarding \textit{AUC} scores both models achieved excellent prediction performances. This is in agreement with the hypothesis that supervised gene selection should improve the classification accuracy. Moreover, the relative performance of the methods is basically the same: RM and PCA method have similar performances. The average value of meta-parameter ($K^{\ast}$) is lower when supervised gene selection is used.

\begin{table}[!h]
\begin{threeparttable}
\caption{Prediction performances of the RM- and PCA-based prediction models using supervised gene selection on the leukemia data set with $36/36$ split of samples over $100$ randomization trials.}
\label{tab:table2}
\begin{tabular}{rccccccc}
\hline
& \multicolumn{3}{c}{RM-LDA} & & \multicolumn{3}{c}{PCA-LDA} \\
\cline{2-4}
\cline{6-8}
\multicolumn{1}{c}{$p^{\ast}$} & \textit{MER} & $K^{\ast}$ & \textit{AUC} & & \textit{MER} & $K^{\ast}$ & \textit{AUC} \\
\hline
$50$ & $0.04\,(0.03)$ & $2.27\,(0.65)$ & $0.99$ & & $0.03\,(0.02)$ & $1.07\,(0.33)$ & $1.00$ \\
$100$ & $0.04\,(0.03)$ & $2.55\,(0.83)$ & $0.99$ & & $0.03\,(0.02)$ & $1.11\,(0.63)$ & $0.99$ \\
$200$ & $0.05\,(0.04)$ & $3.07\,(0.96)$ & $0.99$ & & $0.03\,(0.02)$ & $1.11\,(0.40)$ & $0.99$ \\
\hline
\end{tabular}
\begin{tablenotes}[flushleft]
\footnotesize
\setlength{\itemindent}{-3pt}
\item Legend: RM-LDA -- RM-based class prediction; PCA-LDA -- PCA-based class prediction; $p^{\ast}$ -- number of selected genes; \textit{MER} -- mean error rate; $K^{\ast}$ -- estimated number of gene factors (components); \textit{AUC} -- area under the ROC curve.
\end{tablenotes}
\end{threeparttable}
\end{table}

\subsection{Application to Prostate data set}
\label{subsec:prostate}
The second data set used in this study is the Prostate tumor data. The experimental design was the same as for the Leukemia data set described previously (see Subsection~\ref{subsec:leukemia}). For both, random and supervised gene selection approaches, a learning set consists of $n_{L}=36$ samples and a test set consists of $n_{T}=66$ samples. In total $6600$ class predictions were generated by each of the dimension reduction method based on $100$ randomization trials.

\begin{figure}[!h]
\includegraphics[width=\textwidth]{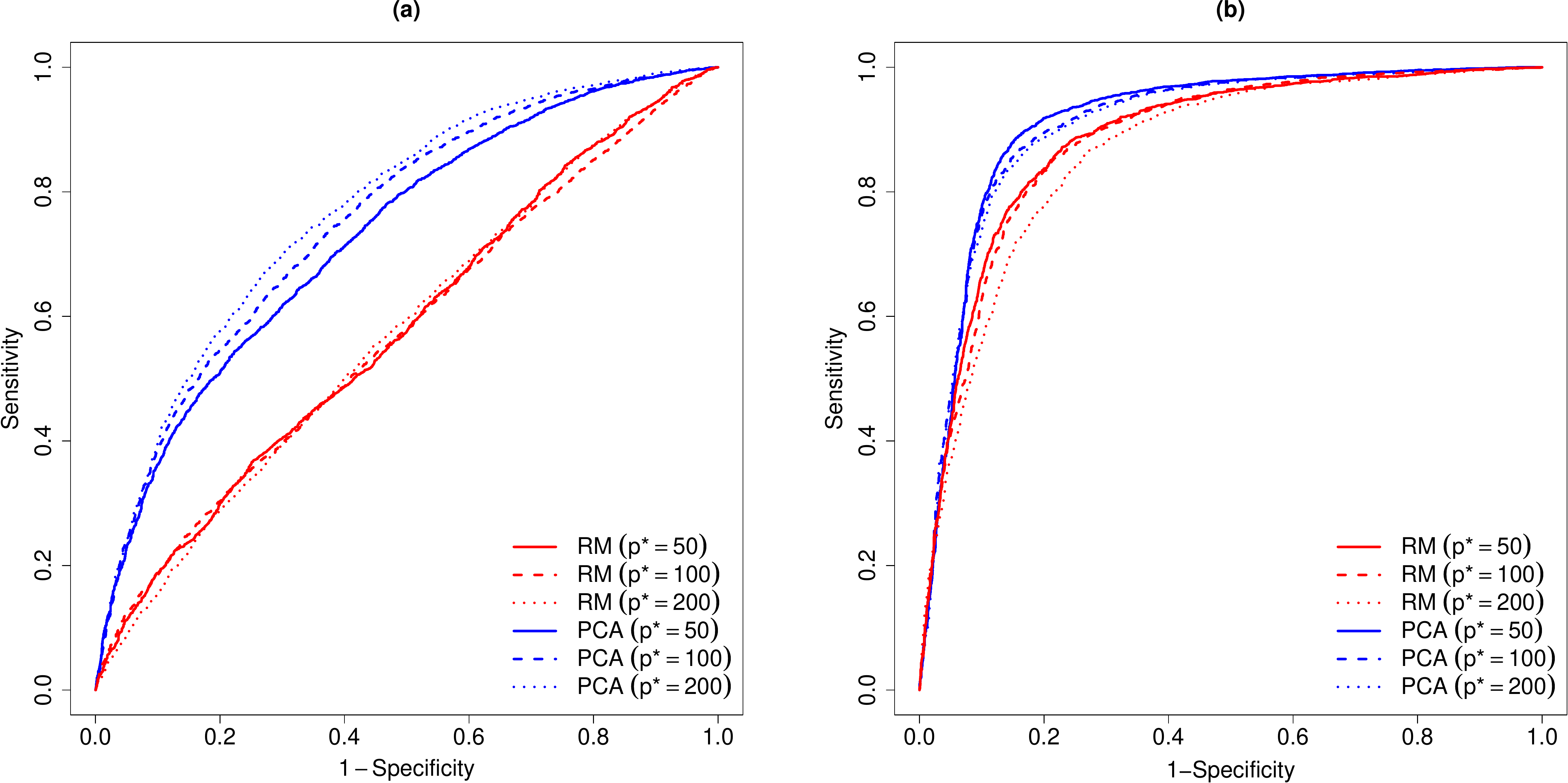}
\caption{Receiver operating characteristic curves (ROC) for RM- and PCA-based prediction models using random (a) and supervised (b) gene selection on the Prostate data set with $36/66$ split of samples over $100$ randomization trials.}
\label{fig:figure4}
\end{figure}

Performance statistics of the RM- and PCA-based prediction models using random gene selection approach are summarized in Table~\ref{tab:table3}. ROC analysis is visualized in Figure~\ref{fig:figure4}(a). Although the pattern of performances between methods on the Prostate data set is similar to that on Leukemia data set, the results suggest that the classes are less well predicted. \textit{AUC} scores indicate that RM-based prediction performances are close to random prediction, while PCA-based approach is generally acceptable.

\begin{table}[!h]
\begin{threeparttable}
\caption{Prediction performances of the RM- and PCA-based prediction models using random gene selection on the Prostate data set with $36/66$ split of samples over $100$ randomization trials.}
\label{tab:table3}
\begin{tabular}{rccccccc}
\hline
& \multicolumn{3}{c}{RM-LDA} & & \multicolumn{3}{c}{PCA-LDA} \\
\cline{2-4}
\cline{6-8}
\multicolumn{1}{c}{$p^{\ast}$} & \textit{MER} & $K^{\ast}$ & \textit{AUC} & & \textit{MER} & $K^{\ast}$ & \textit{AUC} \\
\hline
$50$ & $0.46\,(0.08)$ & $2.01\,(1.19)$ & $0.58$ & & $0.34\,(0.10)$ & $3.48\,(1.46)$ & $0.73$ \\
$100$ & $0.45\,(0.08)$ & $2.27\,(1.29)$ & $0.57$ & & $0.32\,(0.10)$ & $3.71\,(1.23)$ & $0.75$ \\
$200$ & $0.45\,(0.08)$ & $2.33\,(1.26)$ & $0.57$ & & $0.30\,(0.11)$ & $3.48\,(1.37)$ & $0.77$ \\
\hline
\end{tabular}
\begin{tablenotes}[flushleft]
\footnotesize
\setlength{\itemindent}{-3pt}
\item Legend: RM-LDA -- RM-based class prediction; PCA-LDA -- PCA-based class prediction; $p^{\ast}$ -- number of selected genes; \textit{MER} -- mean error rate; $K^{\ast}$ -- estimated number of gene factors (components); \textit{AUC} -- area under the ROC curve.
\end{tablenotes}
\end{threeparttable}
\end{table}

Application of supervised gene selection using Welch's $t$-test yielded much better prediction performances (Table~\ref{tab:table4} and Figure~\ref{fig:figure4}(b)). \textit{MER}s and \textit{AUC}s increase substantially. Moreover, the differences between both methods are minimal. Regarding ROC analysis both model performs excellent, although the \textit{AUC} scores for the RM-based model are slightly lower.

\begin{table}[!h]
\begin{threeparttable}
\caption{Prediction performances of the RM- and PCA-based prediction models using supervised gene selection on the Prostate data set with $36/66$ split of samples over $100$ randomization trials.}
\label{tab:table4}
\begin{tabular}{rccccccc}
\hline
& \multicolumn{3}{c}{RM-LDA} & & \multicolumn{3}{c}{PCA-LDA} \\
\cline{2-4}
\cline{6-8}
\multicolumn{1}{c}{$p^{\ast}$} & \textit{MER} & $K^{\ast}$ & \textit{AUC} & & \textit{MER} & $K^{\ast}$ & \textit{AUC} \\
\hline
$50$ & $0.18\,(0.08)$ & $2.91\,(1.23)$ & $0.88$ & & $0.14\,(0.06)$ & $2.06\,(1.25)$ & $0.91$ \\
$100$ & $0.19\,(0.07)$ & $3.35\,(1.19)$ & $0.88$ & & $0.15\,(0.06)$ & $2.41\,(1.29)$ & $0.91$ \\
$200$ & $0.21\,(0.08)$ & $3.55\,(1.28)$ & $0.86$ & & $0.15\,(0.07)$ & $2.72\,(1.30)$ & $0.91$ \\
\hline
\end{tabular}
\begin{tablenotes}[flushleft]
\footnotesize
\setlength{\itemindent}{-3pt}
\item Legend: RM-LDA -- RM-based class prediction; PCA-LDA -- PCA-based class prediction; $p^{\ast}$ -- number of selected genes; \textit{MER} -- mean error rate; $K^{\ast}$ -- estimated number of gene factors (components); \textit{AUC} -- area under the ROC curve.
\end{tablenotes}
\end{threeparttable}
\end{table}

\section{Discussion}
\label{sec:discussion}
In this paper we explored the possibility of RMs to solve the course of dimensionality problem arising in the context of microarray gene expression data, and evaluated its performance in class prediction framework using LDA. To our knowledge, this is the first extensive validation study addressing RMs for microarray data analysis. In terms of using RM-based dimension reduction of microarray data, the evaluated approach appears to be as effective as widely used PCA-based dimension reduction.

Theoretically RMs can handle a large number of genes. However, as many other multivariate methods it is challenged by large computational time and danger of over-fitting. Therefore, we have used unsupervised random selection of small subset of genes and supervised Welch's $t$-test gene selection procedure. Applying random selection and using the \textit{MER} and \textit{AUC} scores as class prediction performances, overall average values of $\mathrm{\textit{MER}}=0.29$ ($\mathrm{\textit{AUC}}=0.72$) and $\mathrm{\textit{MER}}=0.45$ ($\mathrm{\textit{AUC}}=0.57$) have been reached for Leukemia and Prostate data sets, respectively. We demonstrated that simple $t$-test improve the prediction performance significantly. Considering supervised gene selection procedure, overall average values of $\mathrm{\textit{MER}}=0.04$ ($\mathrm{\textit{AUC}}=0.99$) and $\mathrm{\textit{MER}}=0.19$ ($\mathrm{\textit{AUC}}=0.87$) have been reached for Leukemia and Prostate data sets, respectively. The patterns of performance measures between RM- and PCA-based procedures were similar, although the results suggested that RMs benefit more from preliminary gene selection. Compared to other studies aimed at class prediction, such as ones by \citet{Boulesteix2004}, \citet*{Dai2006}, or \citet*{Nguyen2004}, our performance values are comparable. Slightly better prediction performances in the case of Leukemia data set confirm the fact that the biological separation between the two classes is more pronounced in Leukemia data set \citep*{Antoniadis2003, DeSmet2004}.

We have developed our approach for discretized microarray data because RM scaling assumes a binary response of a gene expression level. Although our results indicate that the loss of information due to discretization step in our procedure is minimal, it is still the issue, if it is reasonable to consider gene expression discretely. Referring to the work of \citet{Sheng2003}, who demonstrated the effectiveness of the Gibbs sampling to the biclustering of discretized microarray data, we argue that discretization may improve the robustness and generalizability of the prediction algorithm with regard to the high noise level in the microarray data. Following \citet{Hartemink2001} the discretization of continuous gene expression levels is preferred for three reasons: (i) gene transcription occurs in one of a small number of states (low--high, off--low--high, low--medium--high, off--low--medium--high, etc.); (ii) the mechanisms of cellular regulatory networks can be reasonably approximated by primarily qualitative statements describing the relationships between states of genes; (iii) discretization, as a general rule, introduces a measure of robustness against error. It is a worthwhile future project to study the performance of our method on different levels of dicretization, using polytomous latent variable models (e.g., partial credit model) that could model more than two states of gene expression.

The major dilemma coupled with class prediction studies is the measurement of the performance of the classifier. The classifier cannot be evaluated accurately when sample size is very low. Moreover, feature selection and feature extraction steps should be an integral part of the classifier, and as such they must be a part of the evaluation procedure that is used to estimate the prediction performance \citep*{Asyali2006}. \citet{Simon2003} reported several studies published in high impact factor journals where this issue is overlooked, and biased prediction performances are reported. To address these issues we applied LOOCV scheme to estimate the appropriate number of gene factors and re-randomization experimental design to stabilize performance measures. It seems that LOOCV is appropriate, but a more sophisticated design (e.g., subsampling, bootstrap sampling, $0.632$ estimator, etc.) to determine the number of gene factors could improve the prediction performance of the RM-based approach.

The vast amounts of gene expression data generated in the last decade have significantly reshaped statistical thinking and data analysis. The approach presented here can be also applied to many other problems in computational biology (e.g., single nucleotide polymorphism (SNP) array data analysis) and could be generalized to the other fields of sciences and humanities (e.g., health studies, risk management, financial engineering, etc.). Hence, innovative statistical methods like the one presented in this paper are of great relevance.

\section{Conclusions}
\label{sec:conclusions}
We have proposed a RM-based dimension reduction approach for the class prediction on microarray gene expression data. Our method is designed to address the course of dimensionality and overcome the problem of ``large $p$, small $n$'' so common in microarray data analysis. Experimental results showed that our procedure appears to be as effective as widely used PCA-based dimension reduction method. We demonstrated that binarization of continuous gene expression levels does not affect prediction performance of the classifier. We showed that appropriate gene selection is crucial before dimension reduction is performed. We restricted our approach to the binary prediction problem, but the methodology can be extended to cover multiclass prediction. The application of our method to other prediction problems (e.g., regression, survival analysis) is straightforward. We are currently working on extending this work on other latent variable models (e.g., partial credit model, Rasch-Andrich model, etc.).

\section*{Acknowledgement}
The first author was supported by Junior Research Fellowship granted by Slovenian Research Agency.

\end{document}